\documentclass[conference]{IEEEtran}
\IEEEoverridecommandlockouts
\usepackage{cite}
\usepackage{amsmath,amssymb,amsfonts}
\usepackage{algorithmic}
\usepackage{graphicx}
\usepackage{textcomp}
\usepackage{xcolor}
\def\BibTeX{{\rm B\kern-.05em{\sc i\kern-.025em b}\kern-.08em
    T\kern-.1667em\lower.7ex\hbox{E}\kern-.125emX}}
\begin{document}

\title{Open-set Classification of Common Waveforms Using A Deep Feed-forward Network and Binary Isolation Forest Models\\

\thanks{We wish to acknowledge the US Army Research Laboratory for their funding of this project.}
}

\author{\IEEEauthorblockN{C. Tanner Fredieu$^1$, Anthony Martone$^2$, R. Michael Buehrer$^1$}
\IEEEauthorblockA{\textit{$^1$Wireless@VT, Bradley Department of Electrical and Computer Engineering, Virginia Tech, Blacksburg, VA, USA } \\
\textit{$^2$US Army Research Laboratory, Adelphi, MD, USA}
\\
Email: christianf@vt.edu, anthony.f.martone.civ@mail.mil, rbuehrer@vt.edu}

}

\maketitle

\begin{abstract}
In this paper, we examine the use of a deep multi-layer perceptron architecture to classify received signals as one of seven common waveforms, single carrier (SC), single-carrier frequency division multiple access (SC-FDMA), orthogonal frequency division multiplexing (OFDM), linear frequency modulation (LFM), amplitude modulation (AM), frequency modulation (FM), and phase-coded pulse modulation used in communication and radar networks. Synchronization of the signals is not needed as we assume there is an unknown and uncompensated time and frequency offset. The classifier is open-set meaning it assumes unknown waveforms may appear. Isolation forest (IF) models acting as binary classifiers are used for each known signal class to perform detection of possible unknown signals. This is accomplished using the 32-length feature vector from a dense layer as input to the IF models. The classifier and IF models work together to monitor the spectrum and identify waveforms along with detecting unknown waveforms. Results showed the classifier had 100\% classification rate above 0 dB with an accuracy of 83.2\% and 94.7\% at -10 dB and -5 dB, respectively, with signal impairments present. Results for the IF models showed an overall accuracy of 98\% when detecting known and unknown signals with signal impairments present. IF models were able to reject all unknown signals while signals similar to known signals were able to pass through 2\% of the time due to the contamination rate used during training. Overall, the entire system can classify correctly in an open-set mode with 98\% accuracy at SNR greater than 0 dB.
\end{abstract}


\section{Introduction}
Given the ever-increasing demand for wireless spectrum (especially below 6 GHz), spectrum sharing between communications and radar systems has become an important research topic. Although this allows communication networks to operate in traditional radar bands, it does require advanced methods for enhanced monitoring and management of spectrum to limit new sources of interference. Many of the current solutions include the use of dynamic spectrum access (DSA) [8] and/or cognitive radio [8], [9]. Many approaches and research efforts [5], [8], [10] are already underway to solve the challenges of these techniques such as implementing spectrum sensing, sharing, and management. Most of these solutions have involved the use of machine learning techniques [1], [3]-[8] due to their classification abilities. 

\section{Related Work}
Neural networks, especially deep networks have become the preferred method for classifying different components or entire signals in the wireless community, as well-documented in [1], [3], [6], [7], [17]. The most popular neural network architecture for the classification of communication and radar signals is the convolutional neural network (CNN) [1], [3], [6], [7], [17]. This architecture has demonstrated excellent performance in signal and modulation identification especially in wireless standards classification [3], [7], [17]. The most recent work with waveform and modulation classification using CNNs is detailed in [1]. Kong, Jung, and Koivunen use two CNNs to classify generalized waveform types and modulations with additive white Gaussian noise (AWGN) and multi-path fading present using a Fourier synchrosqueezing transform (FSST) with size 1024 as the input. The classified waveforms included SC-radar, SC, and multi-carrier or OFDM. These models have demonstrated accuracy rates of 100\% for the waveforms with AWGN present and $>$95\% with fading present over a 0 dB to 20 dB SNR range. Like [1], AWGN and fading will be the primary impairments used in this paper. AWGN and fading are the most common signal impairments used in experiments [1], [3], [6], [7] as they have the most effect on model accuracy. Normally, AWGN will be used even if no other signal impairments are.

Although, CNNs have been proven time and time again, they are not the only method that can be used. Deep feed-forward and recurrent neural network (RNN) networks have also shown much promise in certain areas [5], but CNNs are still the preferred architecture. CNNs and LSTMs have the benefit of smaller model sizes compared to normal deep neural networks (DNN) at the cost of longer training times being required. 

In this paper, we will examine a different technique also tested in [17] using an effective deep feed-forward model to achieve results similar or better than the state-of-the-art presented in [1] for the classification of seven common waveforms used in radar and communication networks. Signals will contain more signal impairments comprised of AWGN, multi-path fading, frequency offset, phase offset, and IQ imbalance. We also use isolation forest (IF) models in the form of binary classifiers for detection of unknown signals to form an open-set recognition system (OSR). Sets of binary classifiers have shown promise in being able to detect unknown data as presented in [20]. Work has also been done recently in identifying unknown RF signals [18] with different methods, but these have seen limited results. Autoencoders have also have been used extensively as anomaly detectors [12]. This paper will focus on the binary classifiers method since it has not been as well-studied in the RF field to our knowledge.

Open-set classification [19] is the ability of a classification system to learn from a labeled training set to classify known data, but also learn how to identify data that is anomalous to the learned data to reject it as unknown. All fields of machine learning are actively involved in making models have OSR abilities [19] since it is difficult to provide label data of all objects. IF models have been used for such tasks.

IF models [16] are based around using decision trees and random forests to isolate anomalous data as a way of detection. Unsupervised learning is used to learn the known dataset. Known and closely similar data are clustered together to form decision regions. Any data that is classified outside of the decision region is isolated from the clustered forest of known points. This isolation acts as a detection of anomalous data. IF models are trained exclusively on known datasets since it can be difficult to gather datasets on all possible signals. However, if there are known anomalous data within the dataset, the contamination rate can be set equal to that of the percentage of dataset that is anomalous during training.

In this paper, we will examine an effective, DNN model that achieves \textbf{1\% increase over the state-of-the-art model} in [1] in the 0 dB to 5 dB range while matching the accuracy at above 5 dB and a \textbf{5\% to 8\% increase over models} in [17] above 0 dB while also being \textbf{able to detect and reject unknown signals.}

\section{Neural Network and Model Architectures}

\subsection{Classifier Model Layout}
As stated previously, the waveform classifier is created by using a deep feed-forward network architecture outlined in Figure 1. The specific network architecture consists of an input using the magnitude of the 65536-FFT of each signal, dropout layers added for regularization to prevent over-fitting, and fourteen total layers including 6 hidden layers consisting of 2048, 1024, 256, 128, 32, and 16 nodes, respectively. The large number of nodes in the initial hidden layers and large FFT size are needed to gather enough features about a signal to perform open-set classification. Network optimization is performed with an Adamax using default settings. Binary cross-entropy is used for the loss function. Gaussian error linear unit (GELU) activation functions are used for each hidden layer while sigmoid activation is used in the output layer to perform classification.

\begin{figure}[t]
\centerline{\includegraphics[scale=0.835]{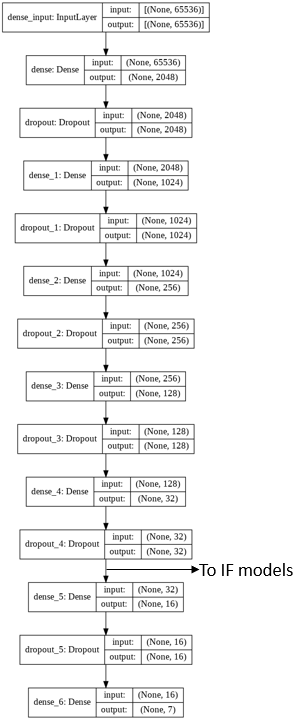}}
\caption{Waveform classifier architecture with the output of the 32-node layer going to IF models}
\label{fig1}
\end{figure}

\subsection{IF Models}
The binary IF models are created using the isolation forest model from the scikit-learn library. Data used from training of the models is outlined in the training section. Contamination was set to 0.02 during each model's training to account for low-noise condition of certain data within the dataset. Detection of anomalous signals follows that all binary classifiers must agree that a signal is known for that signal to be considered known. If any of the classifiers do not agree, the signal is considered unknown. This process is outlined in the flowgraph in Figure 2. In total, seven IF models were created for each known waveform signal class.

\begin{figure}[t]
\centerline{\includegraphics[scale=0.32]{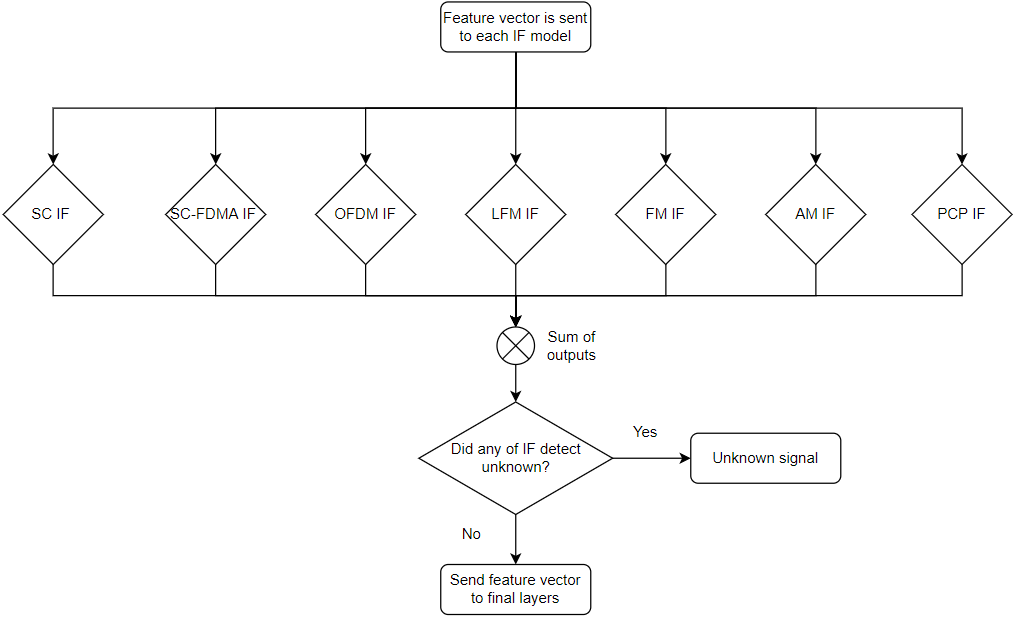}}
\caption{Flowgraph of IF anomaly detection process}
\label{fig2}
\end{figure}

\subsection{Optimization and Regularization}\label{AA}
Regularization, a method to prevent machine learning models from over-fitting on data during training [14], is performed using different techniques for each model. Dropout is used for the classifier as it proves to be the most effective during optimization. This technique causes certain nodes at each layer to drop randomly from use to reduce the co-adaptions formed from backpropagation during training [15]. 2$^n$ number of possible reduced neural networks are formed sharing the weights where n is the number of nodes, and an approximate averaging method is used to simplify all the trained networks into a single one [15].

L2 regularization was found to be optimal with the anomaly detector. L2 regularization, also referred to as weight decay, uses a penalty term added to the cost function to prevent the values of the network weights from becoming too large [14]. The value of dropout, value of L2, number of nodes, number of layers, and learning rate were all determined after going through multiple optimizations to find the ideal parameters.

The optimizer is used to apply the gradient to the network allowing the network to learn. Adam, Adamax, and SGD optimizers were compared with Adamax showing the highest accuracy performance for both models. Adam optimization, as well as the general concept behind Adamax version of Adam, is documented in [2] which introduced the technique.

\section{Experimental Setup}
In this section, we will outline the datasets used to train and test the models. Each waveform class, known and unknown, is individually normalized using the preprocessing.normalize function from the scikit-learn library for all datasets.
\subsection{Training Data}
The training data consists of 10,000 samples of each waveform class for a total of 70,000 signals. All signals are generated using MATLAB scripts in the R2018b version. Modulation for the communication signals include BPSK, QPSK, 16-PSK, 64-PSK, 4-QAM, 16-QAM, 64-QAM, and 256-QAM. Signal bandwidth frequencies included 25 MHz, 50 MHz, 60 MHz, 75 MHz, 80 MHz, and 100 MHz for all classes. The sampling rate was set to 100 MHz except for AM and FM as the upper limit of the sampling rate for these waveforms required to be twice the frequency. Other variations of the waveforms are also included. For OFDM and SC-FDMA, subcarrier spacing is varied among 15 kHz, 30 kHz, and 60 kHz. Single and double sidebands are used for AM. Frank, Barker, and Zadoff-Chu codings are used for the phase-coded pulse signals. 

Environmental signal impairments were added to the dataset. Impairments include -20 dB to 20 dB AWGN, frequency offset of -5000 Hz to 5000 Hz based on the signals sampling frequency, carrier phase offset of -$\pi$ to $\pi$, IQ imbalance of 0 dB to 3 dB, and Rayleigh and Rician fadings.

\subsection{Testing Data}
The testing dataset consists of 3600 signals with 400 signals belonging to each of the seven signal classes and 800 additional signal waveforms belonging to the unknown class. All signals are generated and impaired in the same manner as the training signals. For unknown signals, we use Bluetooth low energy (BLE) 5.0 and white noise signals not included in the training dataset. Bluetooth signals are 2 MHz bandwidth signals with a 125 MHz sampling rate lasting 10 ms produced by one or more Bluetooth devices. Each signal also contains frequency hopping throughout the signal that can potentially exhibit a higher bandwidth than 2 MHz. White noise signals are sampled at the same rate and frequencies as training data and last for 1 ms. Unknown signals contain all aforementioned signal impairments.

\subsection{Training and Testing Methods}
Training epoch is set to 45 with the batch size set to 128 samples per batch for the classifier and uses all 70,000 training signals for training and validation. 5,000 of each signal class is randomized and used to train each binary IF classifier with contamination set to 0.02. As stated previously, this is performed to keep the model from focusing too much on noise components in some samples. Testing is performed in three phases. In Phase 1, testing of the classifier's closed-set performance is evaluated by using only known signals with comparison to known CNN models. Phase 2 tests the performance of the IF models with known and unknown signals along with a comparison to another similar model used for binary classification such as the support vector machine (SVM). Lastly, in Phase 3, the performance of the whole system is tested.

\section{Results}
In this section, we will show the performance results from the classifier given known signals and the knowledge graph's ability to detect unknown signals.

\subsection{Known Signal Classification}
The overall performance of the classifier tested against signals with all impairments is illustrated in Figure 3. Due to the classifier's ability to correctly identify all classes except SC-FDMA and OFDM even at low SNR values, Figure 3 shows the classification performance across SNR values of each individual waveform type so the full scope of the classifier's abilities can be illustrated. The classifier is able to achieve 100\% across all waveforms beginning at 0 dB and achieve above 85\% accuracy for all waveforms at -5 dB. Most of the error results from classifications of OFDM and SC-FDMA as shown in the confusion matrix illustrated in Figure 7.

Figure 4, Figure 5, and Figure 6 are shown to illustrate the minimal effect the signal impairments have on the classifier's performance. Phase offsets have little effect on classification while frequency offsets have approximately less than 5\% accuracy difference in the lowest SNR value and less than 2\% in other cases. IQ imbalance has no effect on classification rates.

Table 1 illustrates the classifier's performance when compared to other known CNN models used for waveform classification. [1] is considered state-of-the-art while models coming from [17] are legacy models that set the standard for initial performance. We show that our classifier, despite being a simpler deep feed-forward network, is able to outperform the legacy models and perform similar to [1]. Our model is actually able to improve classification accuracy of OFDM signal by approximately 3\% when compared to [1].

\begin{figure}[t]
\centerline{\includegraphics[scale=0.4]{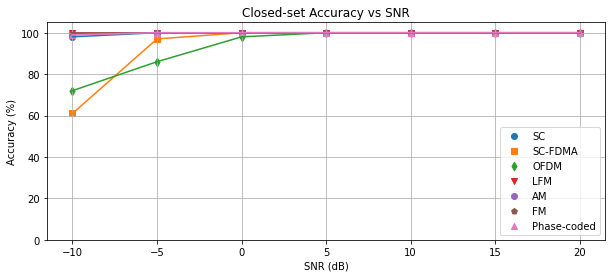}}
\caption{Classifier performance with all impairments}
\label{fig3}
\end{figure}

\begin{figure}[t]
\centerline{\includegraphics[scale=0.32]{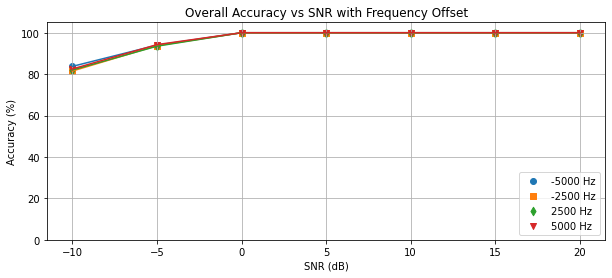}}
\caption{Impact of frequency offset on classifier performance with fading}
\label{fig4}
\end{figure}

\begin{figure}[t]
\centerline{\includegraphics[scale=0.32]{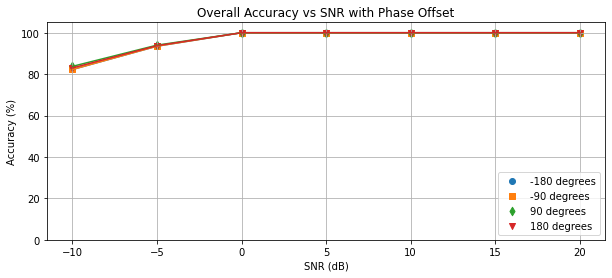}}
\caption{Impact of phase offset on classifier performance with fading}
\label{fig5}
\end{figure}

\begin{figure}[t]
\centerline{\includegraphics[scale=0.32]{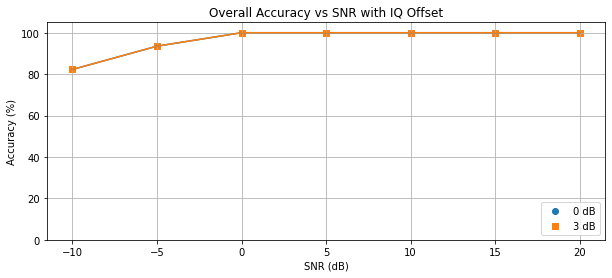}}
\caption{Impact of IQ imbalance on classifier performance with fading}
\label{fig6}
\end{figure}

\begin{figure}[b]
\centerline{\includegraphics[scale=0.38]{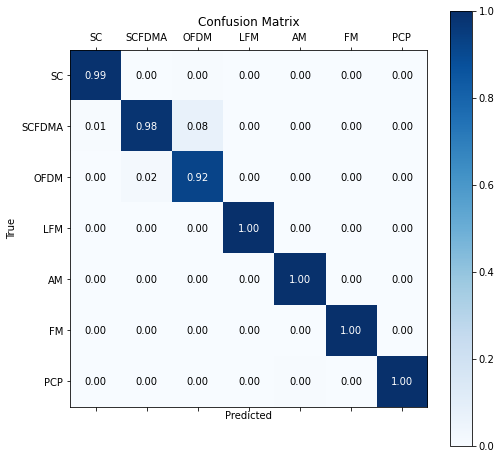}}
\caption{Confusion matrix of classifier performance from -10 dB to 20 dB}
\label{fig7}
\end{figure}

\begin{table}[t]
\caption{Comparison of previous CNNs' performances with classifier's closed-set performance}
\begin{center}
\begin{tabular}{|c|c|c|c|c|c|}
\hline
\cline{2-4} 
\textbf{Model} & \textbf{\textit{0 dB}} & \textbf{\textit{5 dB}} & \textbf{\textit{10 dB}} & \textbf{\textit{15 dB}} & \textbf{\textit{20 dB}}\\
\hline
CNN [1] & 97\% & 99\% & 100\% & 100\% & 100\%\\
\hline
CNN1 0.5 dropout [17] & 90\% & 93\% & 94\% & 94\% & 94\%\\
\hline
CNN2 0.5 dropout [17] & 93\% & 95\% & 95\% & 95\% & 95\%\\
\hline
CNN2 0.6 dropout [17] & 93\% & 95\% & 95\% & 95\% & 95\%\\
\hline
Proposed Classifier & 98\% & 100\% & 100\% & 100\% & 100\%\\
\hline
\hline
\end{tabular}
\label{tab1}
\end{center}
\end{table}

\subsection{Unknown or Anomalous Signal Classification}
For this section of the experiment, all 3600 signals were used. Figure 8 shows the performance accuracy of the binary IF models and binary SVM models when using known and unknown signals. Accuracy is averaged over 7 models. However, due to similar performance, the average accuracy can be interpreted as the accuracy of any individual model. Accuracy is similar between both model types when classification is performed with known signals. SVM is able to correctly identify signals approximately 95\% across all SNR values while IF is able to slightly outperform with approximately 98\% across all SNR values. Almost all errors were due to the similarity between known signals such as SC-FDMA and OFDM causing confusion. However, it is quickly illustrated that the IF models drastically outperform the SVM models when identifying unknown signals. SVM is only able to reach a max rate of 63\% while IF is still able to reach 98\% accuracy.

\begin{figure}[t]
\centerline{\includegraphics[scale=0.4]{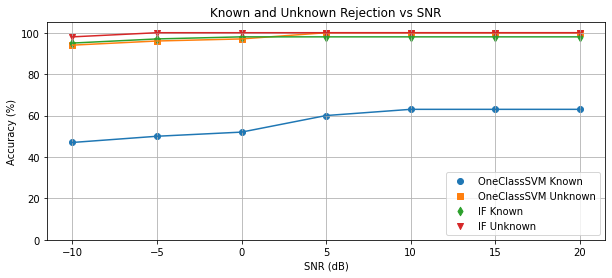}}
\caption{Binary classifiers performance}
\label{fig8}
\end{figure}

\subsection{OSR System}
Accuracy rates for the entirety of the OSR system were as expected. Open-set classification rates remained similar to the closed-set rates. Difference in classification rates come from the IF models when signals that were similar enough to known signals were able to make it through to the classifier. It should be noted that this is still only 2\% of the time which can countered by incorporating a confidence metric or multiple classification. Because of this, the max accuracy rate is now upper bounded by the IF models meaning 98\% is the highest classification rate that can be achieved in open-set settings.

\section{Conclusion}
In this paper, we examined a method comprised of a deep feed-forward network as a waveform classifier that is slightly more effective then state-of-the-art CNN models, and a set of multiple binary classifiers to perform detection of unknown signals (i.e. open-set classification) with an isolation forest architecture. The signals are first converted to baseband, transformed to the frequency domain, and then the magnitude is taken. In a closed-set scenario, the signals are sent to just the classifier for classification as either SC, SC-FDMA, OFDM, LFM, AM, FM, or phase-coded pulse waveform. In an open-set operator, the signals are first sent to the classifier and then the IF models use the 32-length feature vectors of the signals taken from classifier to determine whether they are known or unknown. If the signals are determined to be unknown by all IF models, then they are classified as such. If the signals are determined to be known by any IF models, then they are sent to the remaining layers of the classifier for classification. The classifier then determines if the signal is a SC, SC-FDMA, OFDM, LFM, AM, FM, or phase-coded pulse waveform.


\begin{thebibliography}{00}
\bibitem{b1} G. Kong, M. Jung, and V. Koivunen, "Waveform Classification in Radar-Communications Coexistence Scenarios," GLOBECOM 2020 - 2020 IEEE Global Communications Conference, Taipei, Taiwan, 2020, pp. 1-6.
\bibitem{b2} Kingma and Ba, “Adam: a method for stochastic optimization,” CoRR, 2015, pp. 1-15.
\bibitem{b3} N. E. West and T. O'Shea, "Deep architectures for modulation recognition," 2017 IEEE International Symposium on Dynamic Spectrum Access Networks (DySPAN), Baltimore, MD, USA, 2017, pp. 1-6.
\bibitem{b4} B. Zhao, S. Xiao, H. Lu and J. Liu, "Waveforms classification based on convolutional neural networks," 2017 IEEE 2nd Advanced Information Technology, Electronic and Automation Control Conference (IAEAC), Chongqing, China, 2017, pp. 162-165.
\bibitem{b5} M. Chen, U. Challita, W. Saad, C. Yin and M. Debbah, "Artificial Neural Networks-Based Machine Learning for Wireless Networks: A Tutorial," in IEEE Communications Surveys \& Tutorials, vol. 21, no. 4, pp. 3039-3071, Fourthquarter 2019.
\bibitem{b6} H. Xia et al., "Cellular Signal Identification Using Convolutional Neural Networks: AWGN and Rayleigh Fading Channels," 2019 IEEE International Symposium on Dynamic Spectrum Access Networks (DySPAN), Newark, NJ, USA, 2019, pp. 1-5.
\bibitem{b7} M. H. Alhazmi, M. Alymani, H. Alhazmi, A. Almarhabi, A. Samarkandi and Y. Yao, "5G Signal Identification Using Deep Learning," 2020 29th Wireless and Optical Communications Conference (WOCC), Newark, NJ, USA, 2020, pp. 1-5.
\bibitem{b8} C. Clancy, J. Hecker, E. Stuntebeck and T. O'Shea, "Applications of Machine Learning to Cognitive Radio Networks," in IEEE Wireless Communications, vol. 14, no. 4, pp. 47-52, August 2007.
\bibitem{b9}R. Shafin, L. Liu, V. Chandrasekhar, H. Chen, J. Reed and J. C. Zhang, "Artificial Intelligence-Enabled Cellular Networks: A Critical Path to Beyond-5G and 6G," in IEEE Wireless Communications, vol. 27, no. 2, pp. 212-217, 2020.
\bibitem{b10} Li, Xiaofan, Dong, Fangwei, Zhang, Sha, Guo, Weibin, "A Survey on Deep Learning Techniques in Wireless Signal Recognition," Wireless Communications and Mobile Computing. 2019. 
\bibitem{b11}I. Goodfellow, Y. Bengio, and A. Courville, "Chapter 14: autoencoders," in Deep learning. Cambridge, MA: The MIT Press, pp. 499--523. 2017. 
\bibitem{b12}J. Bui, R. Marks II, "Autoencoder Watchdog Outlier Detection for Classifiers," In Proceedings of the 13th International Conference on Agents and Artificial Intelligence, vol. 2, pp. 990-996. 2021.
\bibitem{b13}G. Kong, M. Jung and V. Koivunen, "Waveform Recognition in Multipath Fading using Autoencoder and CNN with Fourier Synchrosqueezing Transform," 2020 IEEE International Radar Conference (RADAR), Washington, DC, USA. pp. 612-617. 2020.
\bibitem{b14}I. Goodfellow, Y. Bengio, and A. Courville, "Chapter 7: regularization for deep learning," in Deep learning. Cambridge, MA: The MIT Press, pp. 224--270. 2017.
\bibitem{b15}N. Srivastava, G. Hinton, A. Krizhevsky, I. Sutskever, and R. Salakhutdinov," Dropout: a simple way to prevent neural networks from overfitting," in The Journal of Machine Learning Research, vol 15, no. 1, pp. 1919--1958. 2014.
\bibitem{b16} F.T. Liu, K.M. Ting and Z. Zhou, "Isolation Forest," 2008 Eighth IEEE International Conference on Data Mining, 2008, pp. 413-422, doi: 10.1109/ICDM.2008.17.
\bibitem{b17}T.J. O’Shea , J. Corgan, T.C. Clancy, "Convolutional Radio Modulation Recognition Networks," Engineering Applications of Neural Networks. EANN 2016. Communications in Computer and Information Science, vol 629. Springer, Cham.
\bibitem{b18} Y. Shi, K. Davaslioglu, Y.E. Sagduyu, W.C. Headley, M. Fowler and G. Green, "Deep Learning for RF Signal Classification in Unknown and Dynamic Spectrum Environments," 2019 IEEE International Symposium on Dynamic Spectrum Access Networks (DySPAN), pp. 1-10. 2019.
\bibitem{b19} C. Geng, S. Huang and S. Chen, "Recent Advances in Open Set Recognition: A Survey" in IEEE Transactions on Pattern Analysis \& Machine Intelligence, vol. 43, no. 10, pp. 3614-3631, 2021.
\bibitem{b20} L. Shu, H. Xu, and B. Liu, "DOC: Deep Open Classification of Text Documents," Proceedings of 2017 Conference on Empirical Methods in Natural Language Processing, Copenhagen, Denmark. 2017. 
\end{thebibliography}
\end{document}